\documentclass[runningheads]{llncs}

\usepackage{eccv}

\usepackage{eccvabbrv}

\usepackage{graphicx}
\usepackage{booktabs}

\usepackage[accsupp]{axessibility}  

\usepackage[pagebackref,breaklinks,colorlinks,citecolor=eccvblue]{hyperref}

\usepackage{algorithm}
\usepackage{algpseudocode}
\usepackage{multirow}
\usepackage{float}
\usepackage[dvipsnames]{xcolor}

\usepackage{orcidlink}

\begin{document}

\title{Motion Keyframe Interpolation for Any Human Skeleton via Temporally Consistent Point Cloud Sampling and Reconstruction} 

\titlerunning{Motion Keyframe Interpolation for Any Human Skeleton}

\author{Clinton Mo\inst{1} \and
Kun Hu\inst{1,}\thanks{Corresponding author.} \and
Chengjiang Long\inst{2} \and
Dong Yuan\inst{1} \and
Zhiyong Wang\inst{1}}

\authorrunning{C. Mo et al.}

\institute{The University of Sydney, Darlington NSW 2008, Australia\\
\email{clmo6615@uni.sydney.edu.au, \{kun.hu, zhiyong.wang, dong.yuan\}@sydney.edu.au}
\and
Meta Reality Labs, Burlingame, CA, USA 
\\ \email{clong1@meta.com}}

\maketitle

\begin{abstract}
In the character animation field, modern supervised keyframe interpolation models have demonstrated exceptional performance in constructing natural human motions from sparse pose definitions.
As supervised models, large motion datasets are necessary to facilitate the learning process;
however, since motion is represented with fixed hierarchical skeletons, such datasets are incompatible for skeletons outside the datasets' native configurations. 
Consequently, the expected availability of a motion dataset for desired skeletons severely hinders the feasibility of learned interpolation in practice.
To combat this limitation, we propose Point Cloud-based Motion Representation Learning (PC-MRL), an \textit{unsupervised} approach to enabling cross-compatibility between skeletons for motion interpolation learning.
PC-MRL consists of a skeleton obfuscation strategy using \textit{temporal point cloud sampling}, and an \textit{unsupervised skeleton reconstruction} method from point clouds. 
We devise a temporal point-wise K-nearest neighbors loss for unsupervised learning. Moreover, we propose First-frame Offset Quaternion (FOQ) and Rest Pose Augmentation (RPA) strategies to overcome necessary limitations of our unsupervised point cloud-to-skeletal motion process. 
Comprehensive experiments demonstrate the effectiveness of PC-MRL in motion interpolation for desired skeletons without supervision from native datasets.
\end{abstract}

\section{Introduction}
\label{sec:intro}
\begin{figure}
    \centering
    \begin{subfigure}{0.3\linewidth}
        \centering
        \includegraphics[width=0.75\linewidth]{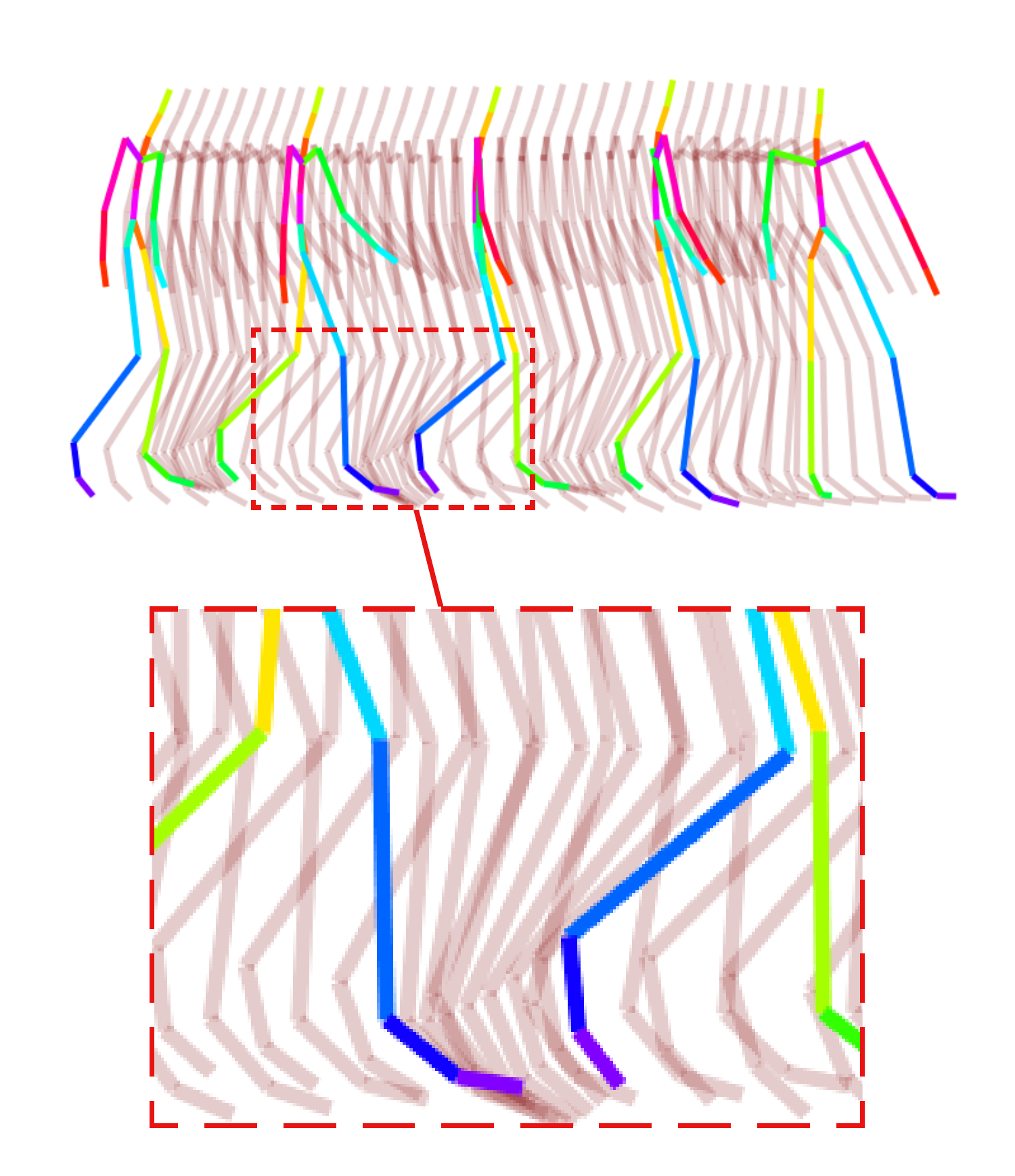}
        \caption{Linear interpolation}
    \end{subfigure}
    \begin{subfigure}{0.3\linewidth}
        \centering
        \includegraphics[width=0.75\linewidth]{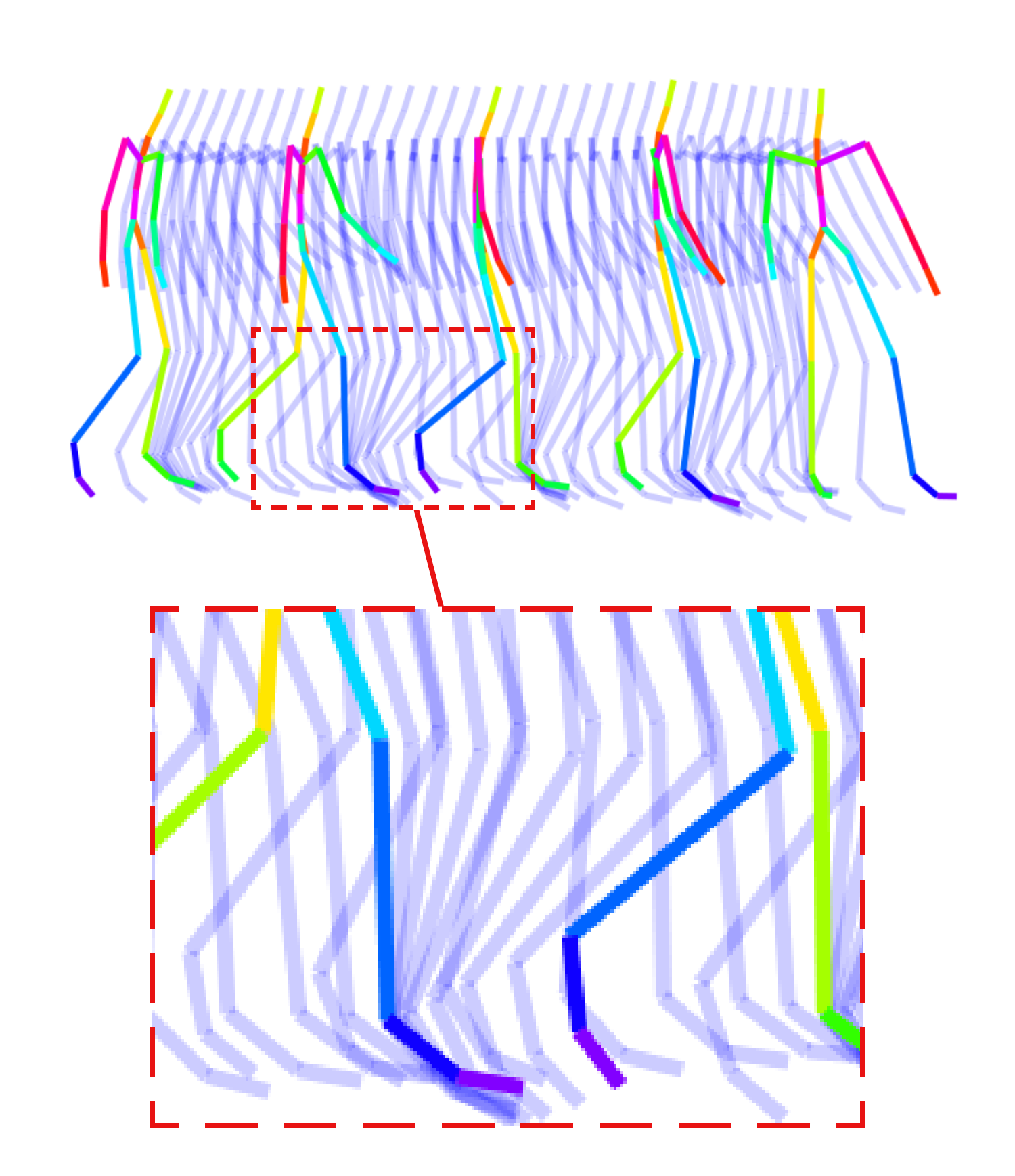}
        \caption{PC-MRL (Our method)}
    \end{subfigure}
    \begin{subfigure}{0.3\linewidth}
        \centering
        \includegraphics[width=0.75\linewidth]{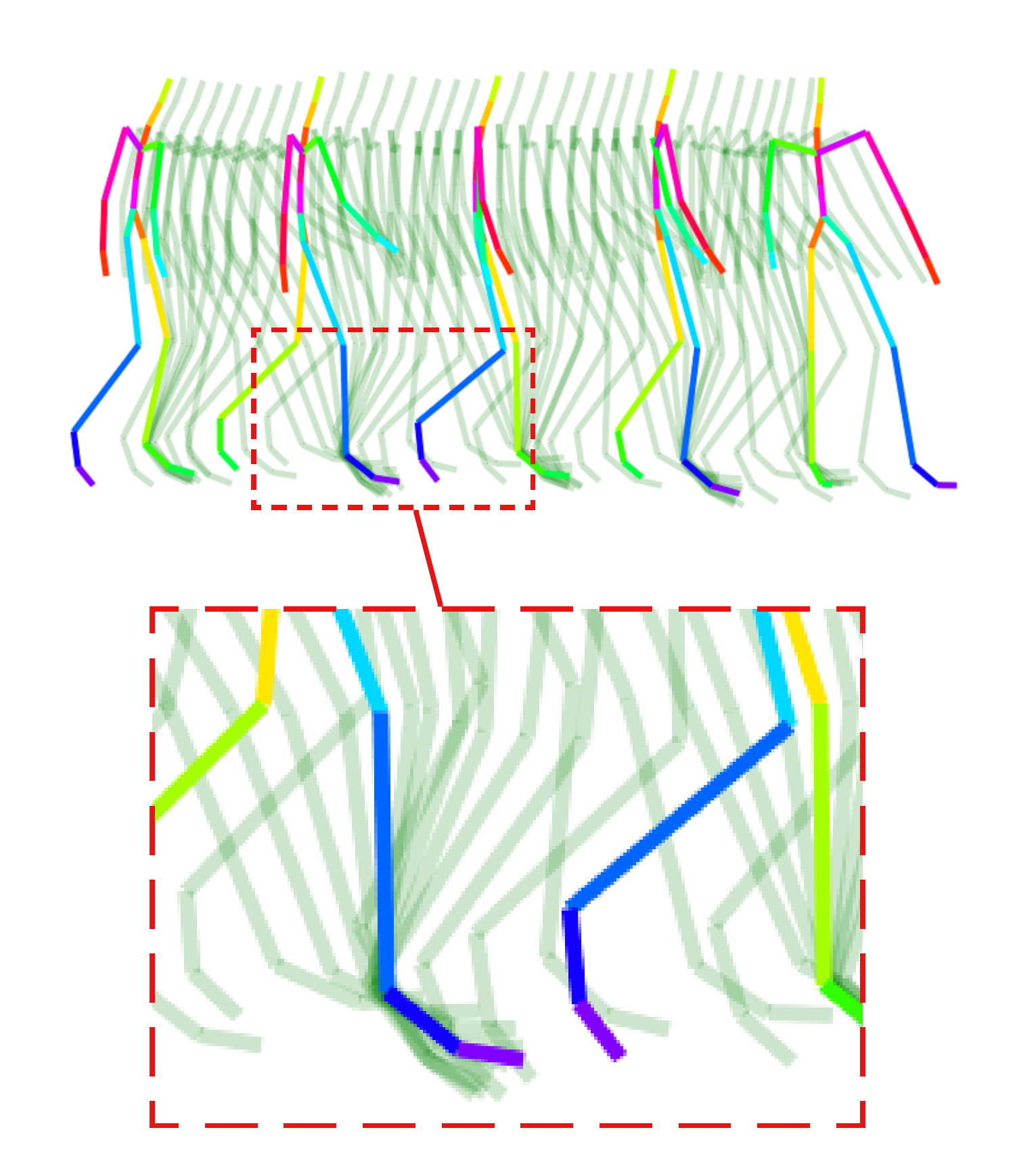}
        \caption{Ground truth}
    \end{subfigure}
    \caption{Motion interpolation of walking keyframes conducted by our approach.}
    \label{fig:interp_intro}
\end{figure}

\begin{figure}
    \centering
    \begin{subfigure}{0.3\linewidth}
        \centering
        \includegraphics[height=3cm]{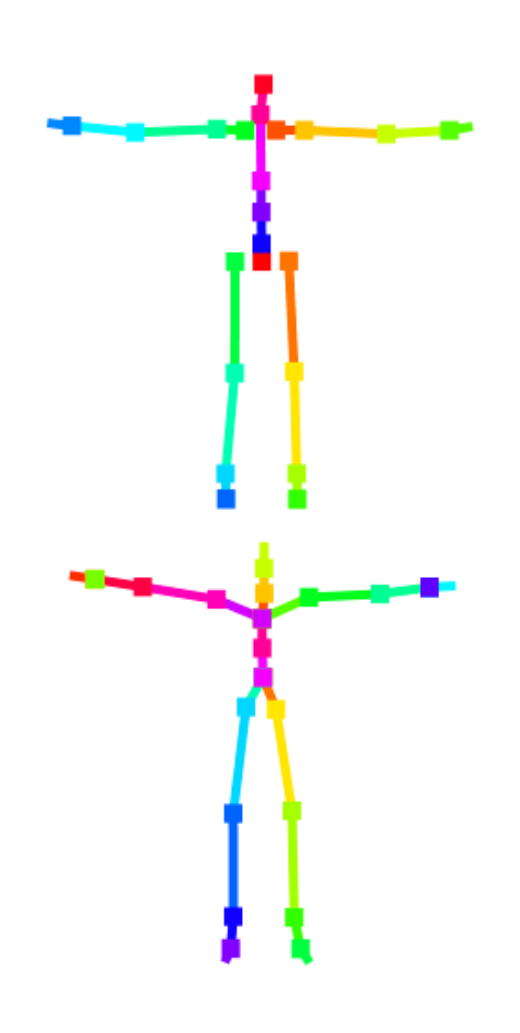}
        \caption{Hierarchical skeleton}
    \end{subfigure}
    \begin{subfigure}{0.3\linewidth}
        \centering
        \includegraphics[height=3cm]{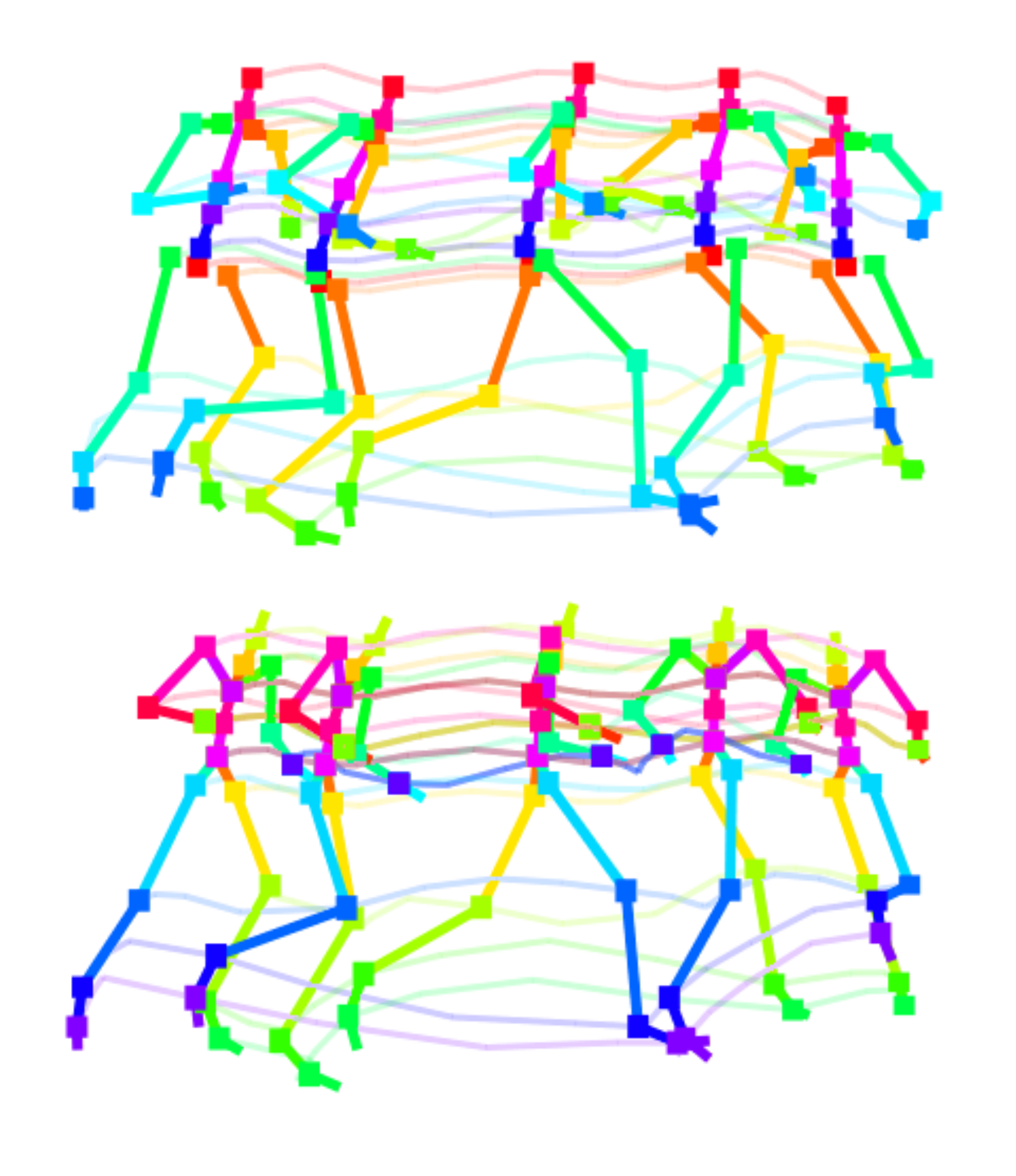}
        \caption{Skeletal motion}
    \end{subfigure}
    \begin{subfigure}{0.3\linewidth}
        \centering
        \includegraphics[height=3cm]{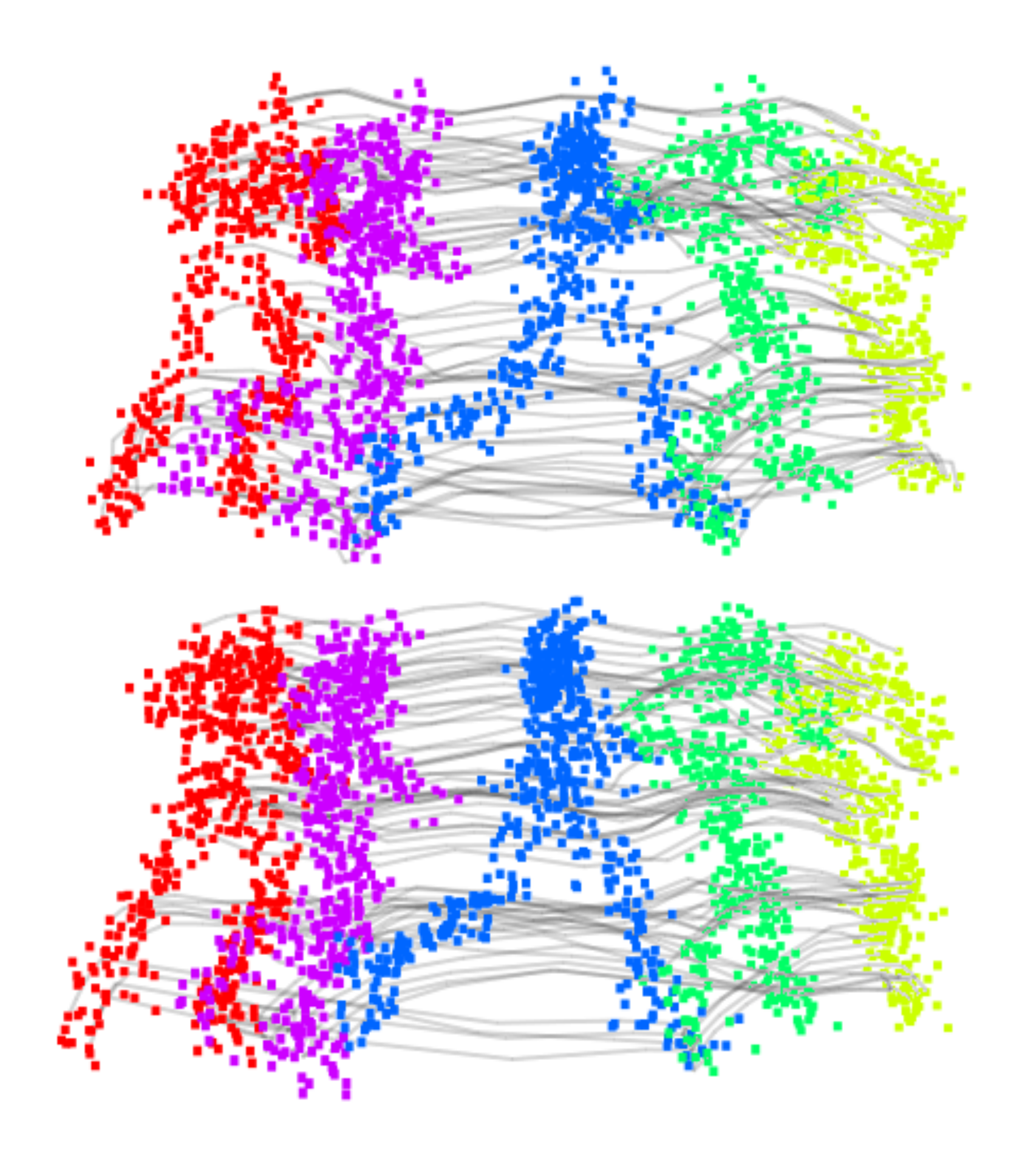}
        \caption{Point cloud representation}
    \end{subfigure}
    \caption{Our method leverages point cloud representations (c) to obfuscate the hierarchical dependencies of skeletal motion. Joint colours correspond to ordinal indices in original motion data representation. Point clouds are coloured by temporal index.}
    \label{fig:point_intro}
\end{figure}

3D character animation workflows largely rely on the concepts of keyframing with interpolation, often referred to as the pose-to-pose principle \cite{frank1981disney}. By defining key poses at correct timings, algorithms can be employed to generate intermediate poses, and thereby eliminating the need for defining each frame individually \cite{reeves1981inbetweening, burtnyk1971computer}. Although this keyframe-based workflow is less costly than frame-by-frame production, human motion is often complex, and natural depictions demand a significant number of keyframes, especially for realistic motions that adhere to physical properties and constraints. 
Although motion capture (MoCap) is frequently employed as an alternative for achieving realistic human motions, MoCap often nonetheless necessitates subsequent keyframing to correct unwanted motion elements and address other imperfections.

In recent years, the interpolation process has been widely studied, particularly with the advent of machine learning methods for sequential data. 
Compared to conventional methods such as linear interpolation (LERP),
machine learning methods have demonstrated the capability to derive visually natural motions from sparser keyframe sets \cite{harvey2020robust, qin2022motion, oreshkin2022motion, mo2023continuous, duan2022unified}, which we can observe in Fig. \ref{fig:interp_intro}.
Data-driven interpolation approaches require large motion capture datasets to learn the necessary motion features for effective synthesis of transitions between keyframes. 
In the field of 3D motions, this aspect significantly restricts the feasibility of learned interpolation methods in animation practice, as motion/pose data are tied to specific skeletal configurations and are not cross-compatible. That is, the dataset's native skeleton, i.e. \textit{source skeleton}, is structurally different and incompatible with the desired skeleton, i.e. \textit{target skeleton}.


Therefore, we propose a novel \textit{point cloud-based motion representation learning} (PC-MRL) approach, in which skeletal hierarchies are obfuscated by sampling into the point cloud medium, and reconstructed into skeletal motion data using an \textit{unsupervised} neural network. 
Unlike raw pose and motion data, the point cloud format is a non-hierarchic representation, and effectively obscures any skeletal configuration as displayed in Fig. \ref{fig:point_intro}.
PC-MRL samples a set of points around the rest position of skeletons using a combination of uniform and normal distributions, and geometrically parents each point to an associated bone in order to effectively represent the associated skeletons' motion data in a temporally consistent manner. 
Subsequently, for the reconstruction of motion data into a given desired skeleton, we propose an \textit{unsupervised} learning scheme, featuring a K-nearest neighbors (KNN) loss between cross-skeleton point clouds to optimize a transformer neural network. 
The network processes the point cloud-based motion representations and generates the corresponding target skeletal motion representations. 
By obscuring both source and predicted target motion data into the point cloud space, our strategy allows for a direct comparison, utilizing KNN to minimize the visual difference between the two point clouds.

The unsupervised nature of our target skeleton reconstruction scheme creates two main disparities between the expected and predicted motion features.
Firstly, the exact roll rotation axis of bones cannot be represented in the point cloud format. To this end, we introduce \textit{First-frame Offset Quaternion} (FOQ) representations to incorporate relative roll values, which are obtainable from \textit{temporally consistent} point clouds. This approach is agnostic to absolute roll rotations, providing a standardisation mechanism for the resulting motion data within the context of motion interpolation learning. 
Second, KNN-based objectives learn skeletal representations in a geometrically optimal manner, which does not always reflect the desired skeletal behaviour. This disparity is commonly observed with smaller skeletal features, such as shoulder and hip bones. As such, we augment training motions using Rest Pose Augmentation (RPA) to increase the range of skeletal behaviours that our interpolation model learns.
Extensive experiments demonstrate the effectiveness of our proposed method, achieving performance targets near the level of direct dataset supervision.

In summary, this paper presents the following key contributions:
\begin{enumerate}
    \item We propose a novel method for achieving unsupervised human motion reconstruction from point cloud data, enabling skeleton-agnostic motion interpolation learning for the first time.
    \item To train motion interpolation models, we formulate first-frame offset quaternions to represent bone rotations with relative roll data, as well as a rest position augmentation strategy to address skeletal configuration variability.
    \item We perform comprehensive experiments to demonstrate our method's effectiveness towards learning motion interpolation, despite the absence of directly compatible datasets in the training process.
\end{enumerate}

\section{Related works}
We explore existing research on motion interpolation approaches, and \textit{motion data modelling} methods in general. In addition, our proposed approach bears resemblance to \textit{unsupervised approaches} for the motion re-targeting research problem, and as such, we will also analyse existing methods in this field.

\subsection{Data-Driven Motion Modelling Methods}

Contemporary research in motion modelling predominantly explore learned data-driven methods, using neural networks to capture, model, and establish correlations between the motion features of various skeletal configurations \cite{mo2021keyframe,zhang2023skinned, zhu2022mocanet, reda2023physics,li2023modular}. 
Motion features alone have demonstrated considerable efficacy in cross-skeleton imitation using a variety of strategies, including task-driven objectives \cite{tiwari2022pose, he2022nemf}, diffusion-based generation \cite{yuan2023physdiff, karunratanakul2023guided, zhang2024finemogen}, and latent feature consistency \cite{aberman2019learning, aberman2020skeleton, villegas2018neural, annabi2024unsupervised, chen2024lart}. 
Latent consistency in particular learns inter-skeleton correlations without the need for paired datasets, which reduces its barrier of entry.

Likewise, motion interpolation strategies have evolved through the use of learning based approaches. The inherent numerical precision of motion data has traditionally posed a challenge for the adoption of the learning approaches, due to their approximate nature \cite{lehrmann2014efficient, hernandez2019human}. This challenge is further amplified by the temporally sparse distribution of keyframes.
Recently, a number of recurrent neural network based methods~\cite{harvey2018recurrent, harvey2020robust, zhang2018data, ren2024diverse} and transformer-based networks~\cite{duan2022unified, mo2023continuous, oreshkin2022motion} have demonstrated encouraging interpolation performance for  the transition between keyframes. 
Additionally, various other motion modelling methods are capable for keyframe based interpolation, albeit with limited intermediate frame representation capabilities \cite{tiwari2022pose, he2022nemf, karunratanakul2023guided}. 

A major constraint inherent in all data-driven motion modelling methods is their dependency on datasets featuring specific skeletal configurations. 
To address this, skeleton-free training strategies have been explored for elementary tasks involving 3D characters. These include vertex-based pose transfer \cite{liao2022skeleton, wang2023zero, chen2023weakly, chen2024lart}, and point cloud-based human shape reconstruction \cite{jiang2019skeleton, wang2020sequential}. 
Notably, in these methods, 3D mesh and point cloud coordinates serve as 
general 3D data representations.
Concurrently, other research has significantly minimized the required volume of training data, bringing it down to single example sequences. This reduction has been achieved through the use of patch matching techniques \cite{li2022ganimator} and imitation learning within physics-based simulations \cite{2021-TOG-AMP, li2024reinforcement}. 
Reinforcement learning has facilitated the generation of goal-driven motion without pre-existing datasets~\cite{2018-TOG-deepMimic, yuan2021simpoe, makoviychuk2021isaac}. However, these simulation-based methods have typically been either exceedingly complex to implement in practice, or yield results that are too inflexible and/or prone to errors for animation workflows.

\subsection{Motion Re-Targeting without Supervision}

Conventional approaches in motion re-targeting have been a cornerstone of 3D animation for decades, primarily dedicated to converting motion capture data into 3D skeletal motions \cite{bodenheimer1997process, gleicher1999animation}. Typically, raw motion capture data is optically recorded, yielding primarily spatial information. In contrast, motion data is largely rotational in nature \cite{burtnyk1971computer}, often represented in \textit{SO(3)} space. Therefore, a principal objective of motion re-targeting solutions has been to effectively bridge these two distinct types of representations.
In~\cite{gleicher1998retargetting}, extensive manually adjustable constraints were introduced for flexible re-targeting between skeletons of identical topology (i.e., same hierarchy, different bone lengths and rest poses). Key practices in this approach, including end-effector matching, joint contacts, and inter-skeleton point correlations, are still relevant to current animation pipelines. 
Point correlations using Inverse Kinematics (IK) have also been proposed as an intermediate representation to facilitate the adaptation between topologically distinct skeletons \cite{monzani2000using, choi2000online}, albeit with hierarchical pairing limitations.

In response, morphologically-independent motion control systems were proposed using IK joint chains and joint-wise constraints~\cite{kulpa2005morphology,hecker2008real}. 
Such systems often expect highly standardised motion skeleton features, and are generally impractical for re-targeting motions produced for different control schemes. 
Additionally, the reliance on Euler angles for extensive classical re-targeting constraints poses a variety of issues regarding rotational freedom \cite{dam1998quaternions}, as well as compatibility with animation workflows \cite{merry2006animation}, which predominantly use quaternion-based systems.

Despite these advancements, the challenge of arbitrary skeleton re-targeting from motion datasets still remains largely unresolved, limiting any widespread application of data-driven pipelines. Our paper aims to effectively address this pervasive limitation.

\section{Methodology}

\begin{figure}[t]
    \centering
    \includegraphics[width=\textwidth]{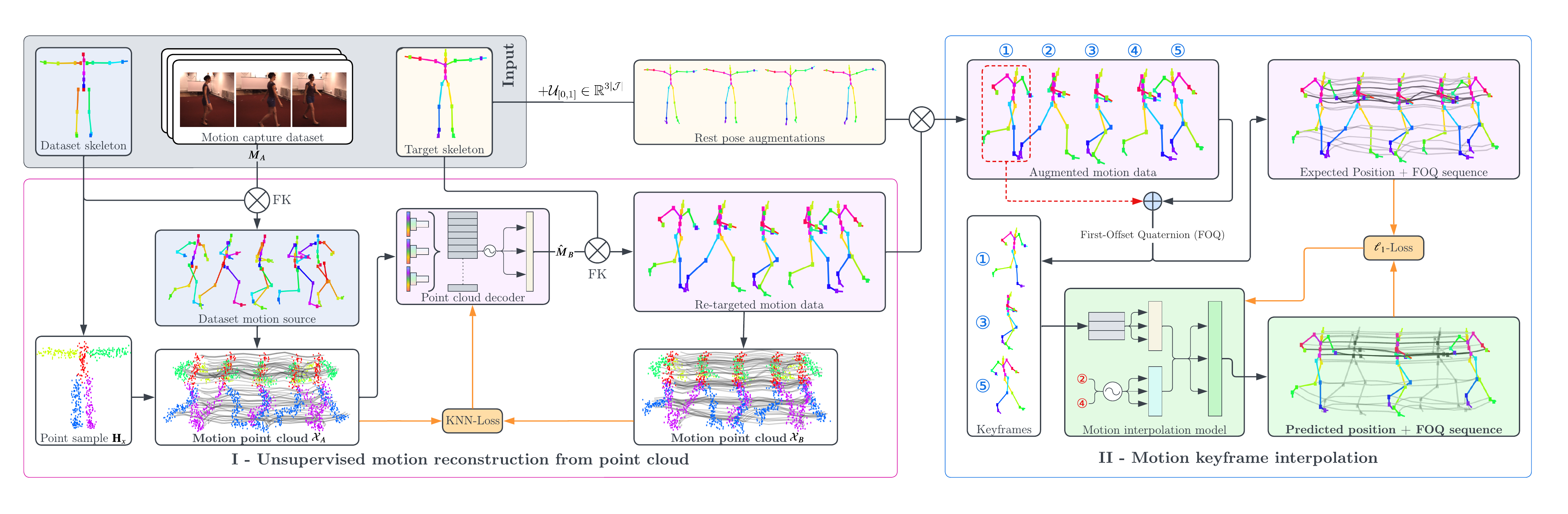}
    \caption{Illustration of point cloud-based motion representation learning (PC-MRL) pipeline for keyframe interpolation to any human skeleton. Our unsupervised point cloud-to-pose decoding approach \textbf{(I)} first adapts motion features from existing datasets to new skeletons. We then employ the adapted motion data to supervise motion interpolation models \textbf{(II)}, using FOQ and rest pose augmentations to address remaining discrepancies between the adapted and expected target skeleton motion patterns. 
    }
    \label{fig:pipeline}
\end{figure}

As illustrated in Fig. \ref{fig:pipeline}, our proposed method consists of two main components: \textit{point cloud-to-motion reconstruction} and \textit{motion keyframe interpolation}. 

\subsection{Quaternion-based Motion Representations}

We first declare a set of notations for representing skeletons. A skeleton $S$ is defined by a tree of bones $\mathcal{B}$. A bone $\mathbf{v}_n \in \mathcal{B}$ is defined by its bone parent $\mathbf{v}_m \in \mathcal{B}$, and a \textit{head position} $\mathbf{H}_n \in \mathbb{R}^3$ which denotes the bone's base offset from its parent head $\mathbf{H}_m$. In addition, a \textit{tail position} $\mathbf{T}_n \in \mathbb{R}^3$ helps to define the end point of the bone's line segment representation, starting from its head position, of $\mathbf{v}_n$. The root bone $\mathbf{v}_0$ has no parent, and thus no head position $\mathbf{H}_0$. The set of $\mathbf{H}_n$ and $\mathbf{T}_n$ for each bone $\mathbf{v}_n \in \mathcal{B}$ make up the \textit{rest position} of the skeleton.

Next, we define the properties of motion data. At any given frame $t$, a skeleton's \textit{pose position} can be represented as global \textit{3D positions} $\mathbf{p}_{n, t} \in \mathbb{R}^3$ and global \textit{quaternion rotations} $\mathbf{q}_{n, t} \in \mathbb{R}^4$ for each bone $\mathbf{v}_n \in \mathcal{B}$. For a given motion sequence $M$, only the root bone is given a 3D position in pose position, i.e. $\mathbf{p}_{0, t}$, as all other pose positions are derived using Forwards Kinematics (FK \cite{dam1998quaternions, kantor1989hypercomplex}):
\begin{equation}
    \label{eqn:fk}
    \mathbf{p}_{n,t} = \mathbf{p}_{m,t} + QR(\mathbf{H}_n, \mathbf{q}_{m,t}),
\end{equation}
where $QR$ is the quaternion rotation function, and $\mathbf{v}_m$ is the parent joint. In summary, a motion sequence $M$ of $|M|$ frames is defined by:
\begin{equation}
    \label{eqn:seq}
    M = \{\mathbf{p}_{0, t} \cup \{\mathbf{q}_{n, t} \mid \mathbf{v}_n \in \mathcal{B}\} \mid 0 \leq t < |M|\}
\end{equation} 

\subsection{Point Cloud Obfuscation and Skeleton Reconstruction}

Our approach is based on the hypothesis that 3D motion sequences can be visually represented by a more universal format: 3D point cloud sequences. 
By geometrically sampling the volume around a skeleton, point cloud data becomes a non-hierarchical representation of its pose(s), obfuscating the original skeletal configuration.
Due to this decoupling between the skeleton configuration and motion representation, a successful motion data reconstruction from point cloud data would be the first step to enabling learnable cross-skeleton motion features.
Formally, for a given source skeleton $S_A$ with an associated motion sequence $M_A$, and any humanoid skeletal configuration $S_B$, the point cloud obfuscation and reconstruction module is to learn a function that can adequately perform 
$F_{S_A\rightarrow S_B}(M_A) = \hat{M}_B$, where $\hat{M}_B$ is a visually similar motion adaptation of $M_A$ on skeleton $S_B$, which is an estimation of $M_B$. 

\subsubsection{Motion Data Obfuscation with Point Clouds}
To sample the point cloud $\mathcal{X}$, we first generate a set of points $u \in \mathcal{X}$ along the line segments defined by the global head and tail positions for each bone $\mathbf{v}_n$. In the bone-local space to $\mathbf{v}_n$, this would mean sampling $u$ by a uniformly distributed factor $\alpha\sim \mathcal{U}_{[0, 1]}$. 
Specifically, $\alpha = 0$ aligns with the head position of the bone, and $\alpha = 1$ indicates the tail position. 
By further sampling $u$ via a normal distribution, our sampling strategy can capture the surrounding volume around the skeleton. Formally, to sample $u$ from its associated bone $\mathbf{v}_n$ within a standard deviation of $\sigma$, we have: 
\begin{align}
    \label{eqn:sample}
    \begin{split}
        \mathbf{H}_x &\sim \mathcal{N}(\alpha\mathbf{T}_n, \sigma), \\
        \mathbf{p}_{u,t} &= \mathbf{p}_{n,t} + QR(\mathbf{H}_x, \mathbf{q}_{n,t})
    \end{split}
\end{align} 
Notably, this sampling process as in Eq. \ref{eqn:sample} is associated with the skeletal configuration only, by which the sampled points are temporally consistent for further processing with the motion sequence.  
Specifically, each point maintains a constant position from its associated bone in the bone-local space. 
The lateral distance of each point from their parent bone segment allows point clouds to react to the bone's rotational roll axis.

In a pure point cloud representation, it is possible to remove the relationships between bones. However, to clearly distinguish symmetrical body features in different skeletons, we introduce and embed body group associations for the points during our reconstruction process. 
Specifically, every bone is categorised by one of these body groups: \textit{[spine, left arm, right arm, left leg, right leg]}. Each sampled point is assigned q group attribute in line with the associated bone. The attribute is represented by a one-hot feature vector $\mathbf{g}_u$. 
To this end, we characterize $u = \{\mathbf{p}_{u,t}, \mathbf{g}_u\} \in \mathcal{X}$. 

\subsubsection{Skeleton Reconstruction}

Given a sampled point cloud representation $\mathcal{X}_A$ from $S_A$ and $M_A$ as input, we aim to train a motion adaptation function $F_{S_A\rightarrow S_B}$, which is tasked with producing a visually similar skeletal motion $\hat{M}_B$ for $S_B$, using a temporal and set-based neural network. 
As shown in Fig. \ref{fig:pipeline}, to construct $F_{S_A\rightarrow S_B}$, we employ a Point Transformer \cite{zhao2021point} to learn embeddings for the unordered point cloud sets frame-wisely, and a standard temporal transformer to process the sequence and decode $\mathcal{X}_A$ into $\hat{M}_B$. 

\begin{figure}[t]
    \centering
    \includegraphics[width=\linewidth]{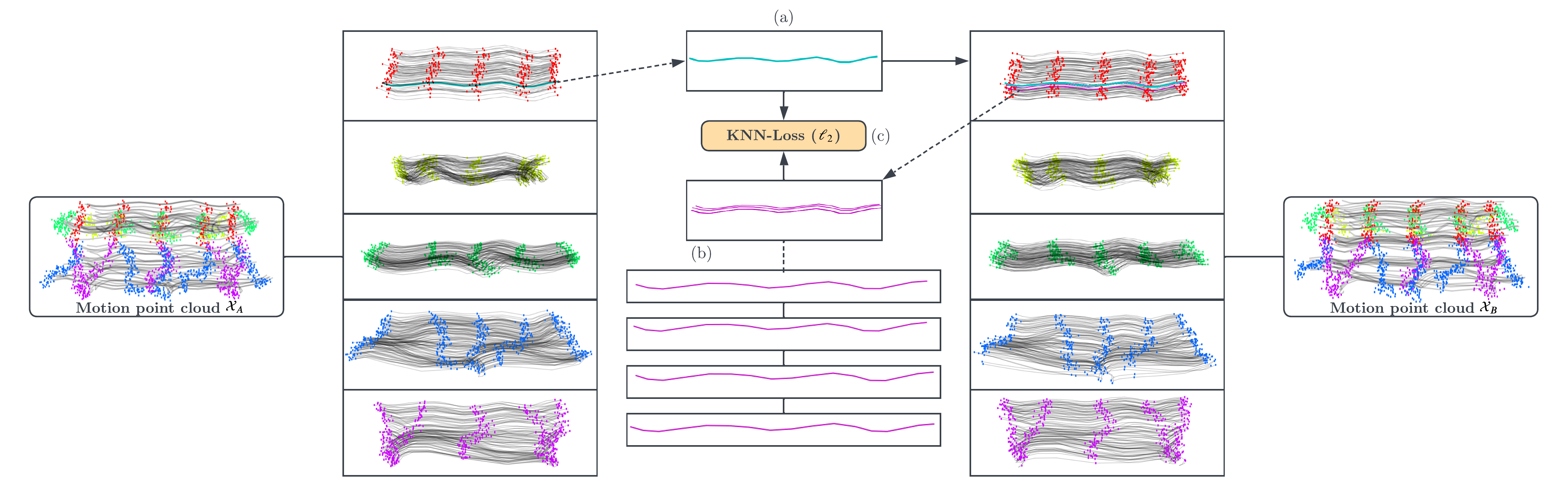}
    \caption{Visualisation of our temporally consistent KNN objective, processed independently per body group. For each point from $\mathcal{X}_A$ (a), the total $\ell_2$ distance throughout the sequence is measured between each point (b) from $\mathcal{X}_B$. The $\ell_2$ distances from the K-nearest points are summed to produce KNN-Loss (c).}
    \label{fig:knnloss}
\end{figure}

To optimize $F_{S_A\rightarrow S_B}$, our objective function maximises the similarity between the sampled point clouds of both $M_A$ and $\hat{M}_B$. 
$\mathcal{X}_A$ can be viewed as the ground truth to $\mathcal{X}_B$, which enables an unsupervised strategy for $F_{S_A\rightarrow S_B}$. 
In detail, we first sample a point cloud $\mathcal{X}_B$ from $S_B$ based on $\hat{M}_B$ derived by $F_{S_A\rightarrow S_B}$. 
Next, a \textbf{temporally consistent KNN-Loss objective} is devised to minimise the $\ell_2$ distance between any given point of $\mathcal{X}_B$ from its $K$ nearest 
neighbouring points in $\mathcal{X}_A$.
The $K$ nearest neighbours are determined based on inter-point distance throughout the entire motion sequence, instead of a frame-wise setting, as shown in Fig. \ref{fig:knnloss}. 
The points with differing body groups are excluded from the neighbour search. Mathematically, we have:
\begin{align}
    \label{eqn:knn}
    &\delta(u_A, u_B) = 
    \begin{cases}
        \sum_{t}||\mathbf{p}_{u_A, t} - \mathbf{p}_{u_B, t}||_2, & \text{if } \mathbf{g}_{u_A} = \mathbf{g}_{u_B},\\
        \inf, & \text{otherwise},
    \end{cases}\\
    &\mathcal{L}_{\text{KNN}}(\mathcal{X}_A, \mathcal{X}_B) = 
    \sum_{u_A\in\mathcal{X}_A, u_B\in N_k(u_A, \mathcal{X}_B)} \delta(u_A, u_B),
\end{align}
where $N_k(u_A, \mathcal{X}_B)$ is the set containing points that are the k-nearest neighbors of $u_A$ in $\mathcal{X}_B$, based on $\delta(u_A, u_B)$ distance. 

In addition to the KNN-Loss, we introduce an optional end-effector loss in the skeleton space for the hand, foot, and head bones, which provides direct positional guidance for matched end-effector pairs $E_{AB}$:
\begin{align}
    \label{eqn:endbone}
    \mathcal{L}_\text{end} &= \sum_{(\mathbf{v}_{A}, \mathbf{v}_{B})\in E_{AB}} \frac{1}{|M_A|} \sum_{t} ||\mathbf{p}_{\mathbf{v}_A, t} - \mathbf{p}_{\mathbf{v}_B, t}||_2.
\end{align}

\subsection{First-Frame Offset Quaternions}

Point clouds lack absolute roll axis information for each bone. As such, we introduce a First-frame Offset Quaternion (FOQ) transform to incorporate the available \textit{relative roll} data for motion modelling. 
We formulate all quaternions of a motion sequence relative to their values from the initial frame.
In detail, for any quaternion sequence $Q = \{\mathbf{q}_0, \mathbf{q}_1, \mathbf{q}_2, ...\}$, we can simply obtain its FOQ transform as $\{\mathbf{q}_0 \times \mathbf{q}_0^{-1}, \mathbf{q}_1 \times \mathbf{q}_0^{-1}, \mathbf{q}_2 \times \mathbf{q}_0^{-1}, ...\}$, where $\mathbf{q}_0^{-1}$ is the quaternion conjugate of $\mathbf{q}_0$.
An FOQ sequence can easily be converted back to a quaternion sequence as long as any $\mathbf{q}_t \in Q$ is known.

Crucially, FOQ is a \textbf{roll-invariant} rotation representation, that is, FOQ remains identical regardless of a bone's absolute roll position in the original data. 
Trivially, to adjust the rotational roll of a bone $\mathbf{v}_n$, we perform the quaternion multiplication for each frame: $\mathbf{q}_{n, t} \times \mathbf{r}$, where $\mathbf{r}$ is a roll quaternion. 
As a quaternion on the roll axis of $\mathbf{v}_n$, $\mathbf{r}$ is constrained within $\mathbf{r} = \alpha + \beta(x\mathbf{i}+y\mathbf{j}+z\mathbf{k})$, where $x$,$y$,$z$ are the 3D XYZ values of $\mathbf{T}_n$, and $\alpha, \beta$ are scalars that control the roll magnitude. 
Note that $\mathbf{r}$ is constant throughout the sequence. Therefore, we can deduce the roll-invariant nature of any constant $\mathbf{r}$: 
\begin{align*}
    (\mathbf{q}_{n, t} \times \mathbf{r}) \times (\mathbf{q}_{0, t} \times \mathbf{r})^{-1} &= \mathbf{q}_{n, t} \times \mathbf{r} \times \mathbf{r}^{-1} \times \mathbf{q}_{0, t}^{-1}\\
    &= \mathbf{q}_{n, t} \mathbf{q}_{0, t}^{-1}
\end{align*}

\subsection{Motion Interpolation Transformer and Rest Pose Augmentation}
\label{sec:model}


To leverage our point cloud-based representation learning for cross-skeleton interpolation, a transformer model - CITL~\cite{mo2023continuous} is adapted, allowing for an efficient interpolation pipeline with existing datasets to alternative skeletons. 
We introduce two key modifications to CITL and its training strategy, focusing on enhancing motion quality.
First, quaternion predictions are substituted with FOQ predictions. This is inspired by the RNN-based method $\text{TG}_{complete}$~\cite{harvey2020robust}, which predicts quaternion offsets from the preceding frame rather than raw quaternions, thereby achieving greater generalisability and superior quality in interpolation. 

For our model's CITL-based architecture, we substitute the original sinusoidal positional encoding scheme with learned relative positions \cite{shaw2018self}. 
While the original implementation relies on the continual nature of sinusoidal functions, we observe that relative positional embeddings with \textit{zero initialisation} converges upon similarly continuous behaviour, 
and significantly lowers the complexity of positional attention. 
The lessened focus on positional relations allows the model to synthesise deeper behaviours within the pose space, improving overall generalisability. As a side effect, the model also accepts arbitrary length inputs beyond the training data length.

Notably, the intended rest pose of a skeletal configuration is often geometrically sub-optimal for point cloud matching, and can manifest in multiple forms.
To address this, we devise a \textbf{rest position augmentation strategy}, leveraging a per-bone Inverse Kinematics (IK) system. RPA broadens the range of intended rest poses that the learned interpolation model can accommodate. 
In detail, during each phase of the Forward Kinematics (FK) process, we augment the bone tails with a random additional offset. 
We re-align the augmented global tail position as closely as possible to its original location using a quaternion rotation with the maximum possible real component. Algorithm \ref{alg:co2q} describes the method by which this quaternion can be derived. Since the real component describes, in cosine, the amount of rotation around an axis, this minimizes numerical disturbance from RPA while altering the visual representation to the desired extent.
\begin{algorithm}[t]
    \caption{IK target alignment with maximum real quaternion component}
    \label{alg:co2q}
    \begin{algorithmic}[1]
        \State \textbf{Input:} $\mathbf{p}_{\text{IK}} = $ 3D IK target unit vector
        \State \hspace{2.7em} $\mathbf{T} = $ resting tail position of associated bone
        \State \textbf{Output:} $\mathbf{q}_{\text{max}} = $ output quaternion, to be applied on original bone quaternion
        \State $\eta \gets \frac{\mathbf{T}}{|\mathbf{T}|}$ \Comment{3D unit coordinates \hspace{0.15em} $[\eta_x, \eta_y, \eta_z]$}
        \State $\omega \gets \frac{\mathbf{p}_{\text{IK}}}{|\mathbf{p}_{\text{IK}}|}$ \Comment{3D unit coordinates $[\omega_x, \omega_y, \omega_z]$}
        \State $\mathbf{q}_{\text{min}} \gets \frac{0 + (\eta_x + \omega_x)\mathbf{i} + (\eta_y + \omega_y)\mathbf{j} + (\eta_z + \omega_z)\mathbf{k}}{2}$
        \State $\mathbf{q}_{\text{max}} \gets \mathbf{q}_{\text{min}} \times (0 + \eta_x\mathbf{i} + \eta_y\mathbf{j} + \eta_z\mathbf{k})$
    \end{algorithmic}
\end{algorithm}

\section{Experimental Results}
\label{sec:results}

\subsection{Datasets}
To thoroughly evaluate the effectiveness of the proposed method, we conducted extensive experiments on the following widely used motion capture datasets:
\begin{itemize}
    \item \textbf{LaFAN1} \cite{harvey2020robust} contains long, high quality motions in controllable video game character styles. The dataset was recorded on a large motion capture stage, and manually cleaned to production standards. The native skeleton of LaFAN1 contains 6 spinal bones, and 4 bones for each limb. 
    \item \textbf{Human3.6M} \cite{h36m_pami, IonescuSminchisescu11} comprises of 3.6 million motion frames.
    The motions captured in Human3.6M tend to be less dynamic, as they were recorded on a relatively compact 4x3m stage. Its native skeleton includes 5 spinal bones, 4 bones for each limb, and an additional finger bone for each hand.
    \item \textbf{CMU Mocap} \cite{cmu} is a diverse motion dataset encompassing 113 categories. It features a mix of static and dynamic motions, captured on a 3x8m stage. The skeleton is identical to that of Human3.6M, with the exception of the pelvis, which is divided into two hip bones and a lower back bone.
\end{itemize}
For PC-MRL, we designated the CMU MoCap skeleton as $S_B$. Its native dataset is exclusively utilized for testing purposes, while LaFAN1 and Human3.6M are used simultaneously for training.\footnote{The authors Clinton Mo, Kun Hu and Zhiyong Wang are signatories of the dataset licenses and produced all experimental results of the paper. Meta Inc. was not granted access to the datasets.}

\subsection{Implementation Details}


In our experiments, point cloud obfuscation adopted a 256-point setting for the sampling process with $\sigma = 0.05$ and each point was characterized by a 64-element vector. 
To maintain consistency, we ensure that all bone offsets and motion positions are defined in \textbf{metre} units. 
For skeletal reconstruction, 4 Point Transformer \cite{zhao2021point} layers was utilized to construct the neural network. 
Within each layer, the feature size was doubled, and the size of the point set was downsampled by a factor of 4 through a farthest-first traversal approach, eventually resulting in a singular feature vector. 
In the final step, two linear layers with ReLU activations were utilized to transform the resulted feature vector in a frame-wise manner, producing the output skeletal representation.
During training, we set $K=8$ for the KNN loss to optimize the network. 
Additionally, the use of unit quaternion values in practice relies on modulus division for constraint, a process that can potentially lead to exploding gradients. To address this, we introduced an objective function $\mathcal{L}_q$ that measures the $\ell_2$ norms' difference between the raw quaternion output and the expected unit quaternion.
All experiments were trained and evaluated on a single NVIDIA GeForce RTX 3090 GPU.


\subsection{Motion Interpolation Comparison against Supervised Methods}

To demonstrate the efficacy of our method, we produce and compare our method against the conventional LERP method, as well as three state-of-the-art interpolation models trained on the original CMU MoCap dataset. Specifically, we measure the performance of the RNN-based $\text{TG}_{\text{complete}}$ model \cite{harvey2020robust}, the BERT-based motion interpolation adaptation \cite{duan2022unified}, and a transformer-based encoder-decoder approach \cite{mo2023continuous}. We additionally provide results on a variant of our method trained without our RPA strategy.
The experimental configuration for each state-of-the-art model is identical to their original implementations. 
To measure interpolation accuracy, we employ the standard $\ell_2$ positional distance (L2P), $\ell_2$ quaternion difference (L2Q), and NPSS \cite{gopalakrishnan2019neural} for visual similarity evaluation \cite{harvey2020robust}.

\begin{table}[t]
    \centering
    \newcommand\boldviolet[1]{\textcolor{Mulberry}{\textbf{#1}}}
    \newcommand\boldblue[1]{\textcolor{blue}{\textbf{#1}}}
    \resizebox{\textwidth}{!}{
        \setlength{\tabcolsep}{5pt}
        \begin{tabular}{|c|l|c c c|c c c|c c c|}
            \hline
            \textbf{Motion} & \multirow{2}{*}{\textbf{Method}} & \multicolumn{3}{|c|}{\textbf{L2P}$\downarrow$} & \multicolumn{3}{|c|}{\textbf{L2Q}$\downarrow$} & \multicolumn{3}{|c|}{\textbf{NPSS}$\downarrow$} \\
            \textbf{category} & & 5 & 15 & 30 & 5 & 15 & 30 & 5 & 15 & 30 \\
            \hline
            \multirow{6}{*}{Basketball} & LERP & 0.0787 & 0.4009 & 0.8033 & 0.1588 & 0.5834 & 0.9611 & \boldblue{0.1315} & 0.6120 & 1.8202 \\
            & $\text{TG}_{\text{complete}}$ \cite{harvey2020robust} & 0.1050 & 0.4419 & 0.7104 & 0.1859 & 0.6439 & 0.9303 & 0.1600 & 0.6579 & \boldblue{1.2136} \\
            & BERT \cite{duan2022unified} & 0.0889 & 0.3737 & 0.6865 & 0.1585 & 0.5560 & 0.8992 & 0.1689 & 0.6177 & 1.4587 \\
            & CITL \cite{mo2023continuous} & \boldviolet{0.0625} & \boldviolet{0.2973} & \boldviolet{0.5708} & \boldviolet{0.1410} & \boldviolet{0.4932} & \boldviolet{0.8011} & \boldviolet{0.1047} & \boldviolet{0.4098} & \boldviolet{1.0959} \\
            \cline{2-11}
            & PC-MRL w/o RPA & 0.0767 & 0.3534 & 0.6715 & \boldblue{0.1505} & \boldblue{0.5431} & \boldblue{0.9049} & 0.1364 & 0.5596 & 1.3471 \\
            & PC-MRL (Ours) & \boldblue{0.0752} & \boldblue{0.3443} & \boldblue{0.6662} & 0.1561 & 0.5495 & 0.9088 & 0.1397 & \boldblue{0.5572} & 1.3015 \\
            \hline
            \hline
            \multirow{6}{*}{Golf} & LERP & \boldviolet{0.0189} & 0.1229 & 0.3187 & \boldviolet{0.0455} & \boldblue{0.1870} & 0.3987 & \boldviolet{0.0587} & 0.3979 & 1.0312 \\
            & $\text{TG}_{\text{complete}}$ \cite{harvey2020robust} & 0.0473 & 0.2616 & 0.4546 & 0.0693 & 0.3191 & 0.5297 & 0.0954 & 0.5062 & 1.1534 \\
            & BERT \cite{duan2022unified} & 0.0276 & 0.1159 & 0.2967 & 0.0612 & 0.2082 & 0.4258 & 0.1099 & 0.3906 & 0.9255 \\
            & CITL \cite{mo2023continuous} & \boldblue{0.0201} & \boldviolet{0.1043} & \boldviolet{0.2589} & \boldblue{0.0476} & \boldviolet{0.1766} & \boldviolet{0.3552} & \boldblue{0.0636} & \boldviolet{0.2581} & \boldviolet{0.7799} \\
            \cline{2-11}
            & PC-MRL w/out RPA & 0.0274 & 0.1304 & 0.3622 & 0.0555 & 0.2065 & 0.4923 & 0.0911 & 0.3759 & 1.2083 \\
            & PC-MRL (Ours) & 0.0273 & \boldblue{0.1130} & \boldblue{0.2706} & 0.0598 & 0.2012 & \boldblue{0.3844} & 0.1049 & \boldblue{0.3716} & \boldblue{0.8150} \\
            \hline
            \hline
            \multirow{6}{*}{Swimming} & LERP & \boldblue{0.0731} & \boldblue{0.3680} & \boldblue{0.7374} & \boldviolet{0.1326} & \boldviolet{0.5327} & \boldblue{0.9476} & 0.2147 & 0.9222 & 1.6972 \\
            & $\text{TG}_{\text{complete}}$ \cite{harvey2020robust} & 0.1292 & 0.5796 & 0.9141 & 0.1845 & 0.7420 & 1.1199 & 0.2660 & 0.9716 & 1.7698 \\
            & BERT \cite{duan2022unified} & 0.1088 & 0.3896 & 0.7516 & 0.1819 & 0.5518 & 0.9622 & 0.2877 & 0.9074 & 1.6682 \\
            & CITL \cite{mo2023continuous} & 0.0905 & 0.3895 & 0.7475 & 0.1484 & 0.5535 & 0.9533 & 0.2084 & \boldviolet{0.8317} & \boldviolet{1.6122} \\
            \cline{2-11}
            & PC-MRL w/o RPA & \boldviolet{0.0722} & 0.3754 & 0.7459 & \boldblue{0.1345} & 0.5442 & 0.9594 & \boldviolet{0.1944} & 0.8731 & 1.6608 \\
            & PC-MRL (Ours) & 0.0734 & \boldviolet{0.3595} & \boldviolet{0.7139} & 0.1355 & \boldblue{0.5331} & \boldviolet{0.9248} & \boldblue{0.1956} & \boldblue{0.8591} & \boldblue{1.6370} \\
            \hline
            \hline
            \multirow{6}{*}{\parbox{2cm}{\centering Walking \& Running Locomotion}} & LERP & 0.0455 & 0.2505 & 0.5527 & 0.1120 & 0.3997 & 0.6233 & 0.0655 & 0.2682 & 0.9071 \\
            & $\text{TG}_{\text{complete}}$ \cite{harvey2020robust} & 0.0540 & 0.1786 & 0.3103 & 0.1147 & 0.2866 & \boldblue{0.3866} & 0.0895 & 0.2616 & \boldblue{0.4265} \\
            & BERT \cite{duan2022unified} & 0.0552 & 0.1998 & 0.3544 & 0.1111 & 0.3101 & 0.4762 & 0.0901 & 0.2760 & 0.6458 \\
            & CITL \cite{mo2023continuous} & \boldviolet{0.0328} & \boldviolet{0.1034} & \boldviolet{0.2427} & \boldviolet{0.0942} & \boldviolet{0.2102} & \boldviolet{0.3384} & \boldviolet{0.0526} & \boldviolet{0.1568} & \boldviolet{0.4080} \\
            \cline{2-11}
            & PC-MRL w/o RPA & 0.0397 & 0.1657 & 0.3417 & \boldblue{0.1058} & 0.2879 & 0.4722 & \boldblue{0.0640} & 0.2288 & 0.6021 \\
            & PC-MRL (Ours) & \boldblue{0.0377} & \boldblue{0.1392} & \boldblue{0.2848} & 0.1064 & \boldblue{0.2778} & 0.4281 & 0.0658 & \boldblue{0.2129} & 0.5098 \\
            \hline
        \end{tabular}
    }
    \caption{Performance comparisons for various motion categories, measured in L2P, L2Q, and NPSS. All methods \textit{except PC-MRL} are trained on the original CMU MoCap dataset. Each motion contains 128 frames with keyframes placed every 5, 15, or 30 frames. The top 2 results for each test are highlighted in \boldviolet{purple} and \boldblue{blue} respectively.}
    \label{tab:interp}
\end{table}

\begin{figure}
    \centering
    \includegraphics[width=0.95\linewidth]{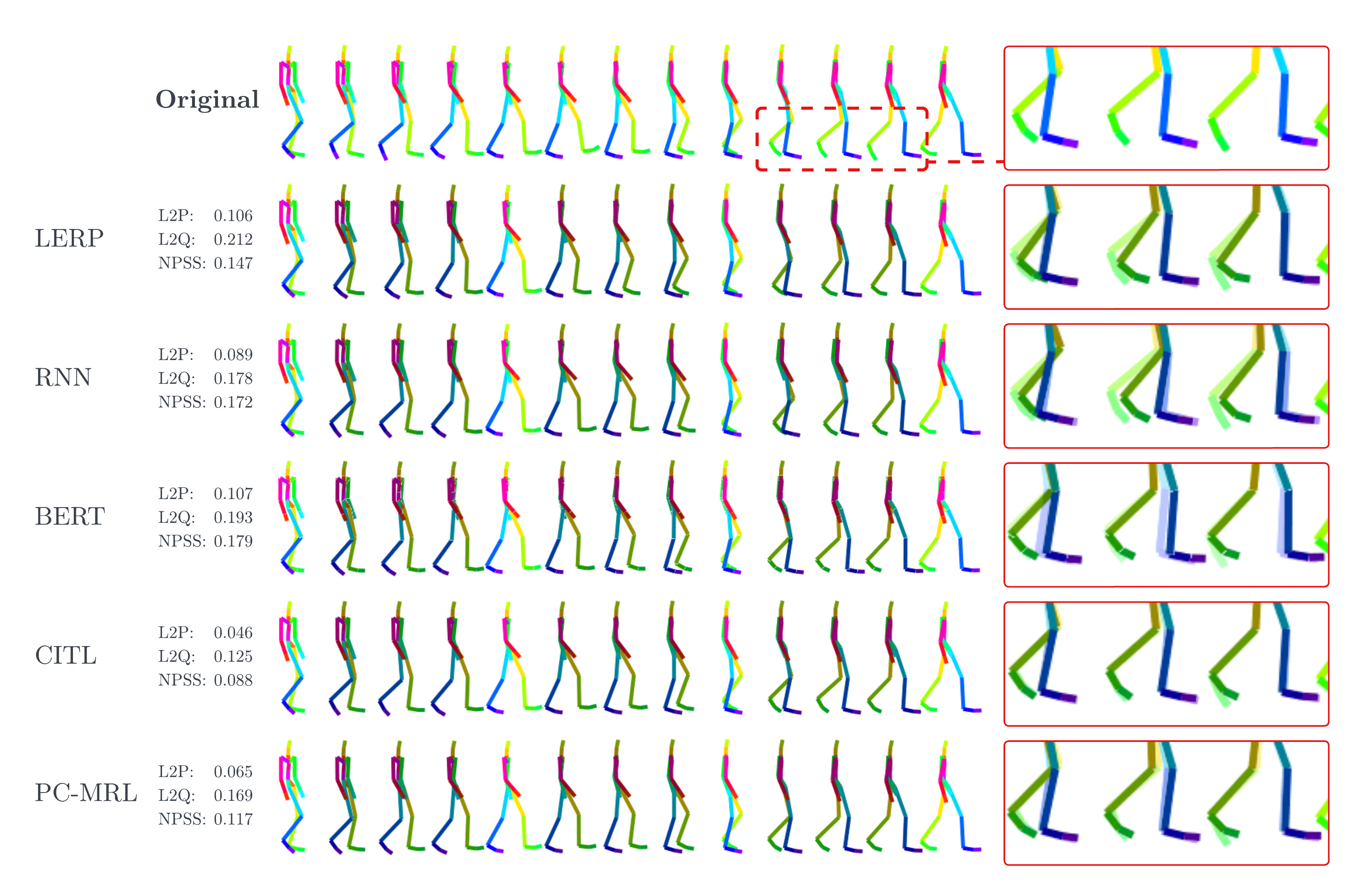}
    \caption{Walking keyframe interpolation examples produced by methods from Table \ref{tab:interp}. Close-up layered comparisons against the original motion are provided on the right.}
    \label{fig:interp}
\end{figure}

The results listed in Table \ref{tab:interp} clearly demonstrate our method's ability to approach the accuracy levels exhibited by directly supervised state-of-the-art methods. PC-MRL consistently outperforms both the LERP standard and RNN-based $\text{TG}_{\text{complete}}$ model, particularly in longer keyframe interval scenarios where the quality and depth of learned motion features is most critical and impactful. Likewise, the inclusion of RPA also tends to improve the long interval performance of PC-MRL, particularly when precise movements (i.e. golf) or deeper motion features (i.e. locomotion) are expected. Given this, we can observe that RPA definitively improves the overall consistency of the PC-MRL method in scenarios that may be unseen or less effectively represented by our point cloud-to-motion system. 


Native supervision, i.e. with CITL, unsurprisingly produces the most accurate interpolation model, with a notable exception of swimming motions, where our PC-MRL supervision produces even stronger results. We believe the abundance of low crawl motions, a similar motion to swimming in LaFAN1, provides more meaningful supervision over the original CMU dataset, which conversely has few similar motions once swimming motions are filtered out. On the other end of the spectrum, i.e. for high data scenarios such as walking and running locomotion from Fig. \ref{fig:interp}, all models including our PC-MRL approach observe the strongest improvements for learned methods over the naive LERP method. Both cases strongly support our method's capability to supervise highly intricate motion patterns, despite the incompleteness of the point cloud representation and reliance on relative rotations and geometric optimisation assumptions.

\subsection{Cross-Skeleton Motion Re-Targeting Experiments}
\label{sec:retarget}

\begin{table}[b]
    \centering
    \resizebox{0.85\textwidth}{!}{
        \setlength\tabcolsep{5pt}
        \begin{tabular}{|l||c|c|c|c|c|c|}
            \hline
            \hfil\multirow{2}{*}{\textbf{Model}}\hfill & \multicolumn{5}{|c|}{$L_{\text{KNN}}(\mathcal{X}_A, \mathcal{X}_B)\downarrow$} & \multirow{2}{*}{$L_{\text{end}}\downarrow$} \\
            \cline{2-6}
             & $\sigma=0$ & $\sigma=0.01$ & $\sigma=0.05$ & $\sigma=0.1$ & $\sigma=0.2$ &  \\
            \hline
            Original motion & 1.0099 & 2.0234 & 5.7399 & 9.6636 & 16.9223 & 0 \\
            \hline
            Primal skeleton \cite{aberman2020skeleton} & 8.1109 & 8.0353 & 9.4206 & 12.8291 & 20.6332 & 9.6990 \\
            PC-MRL (Ours) & \textbf{3.8714} & \textbf{4.1231} & \textbf{6.7277} & \textbf{10.3078} & \textbf{17.5261} & \textbf{5.6258} \\
            \hline
        \end{tabular}
    }
    \caption{Average unpaired LaFAN1 to CMU motion re-targeting performance of the primal skeleton method and our point cloud method.
    $\mathcal{L}_{\text{KNN}}$ is measured using 1024-point clouds at $K=8$. Independently sampled point clouds from the original LaFAN1 skeleton are provided as optimal $\mathcal{L}_{\text{KNN}}$ reference.
    All values are scaled by $\times 100$.}
    \label{tab:retarget}
\end{table}

We further conduct an experiment on motion re-targeting to directly benchmark our method against the sole existing state-of-the-art method for unpaired motion re-targeting with unrestricted skeletons, i.e. \textit{primal skeletons} (PS) \cite{aberman2020skeleton}. For direct motion re-targeting evaluation, we utilized the KNN-Loss and end-effector loss as metrics, owing to their effective measurement of visual similarity and non-sensitivity towards absolute roll axis correctness. Though absolute rolls are generally necessary for a complete motion re-targeting solution, this aspect is out of scope for our project as our main goal is motion interpolation. 

\begin{figure}[t]
    \centering
    \includegraphics[width=0.95\linewidth]{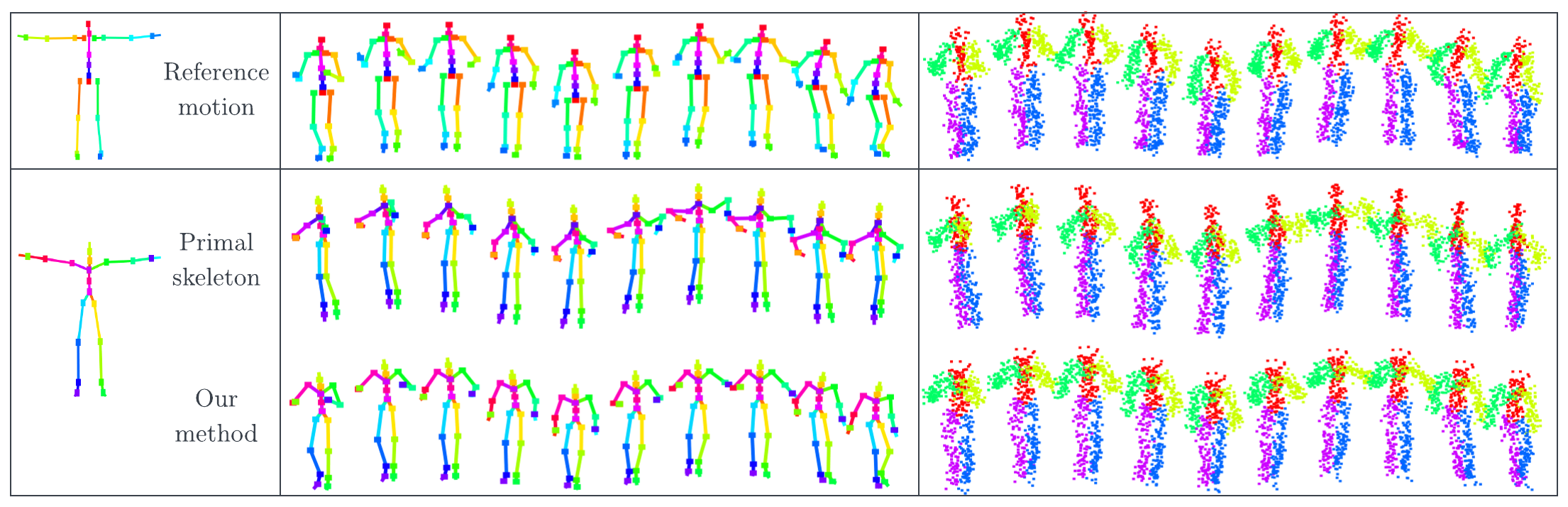}
    \caption{Jumping motion re-targeting examples produced on the CMU skeleton by the primal skeleton method (middle row) and our method (bottom row). Sampled point cloud representations of each motion sequence are provided on the right.}
    \label{fig:retarget}
\end{figure}

\begin{figure}
    \centering
    \includegraphics[width=0.8\linewidth]{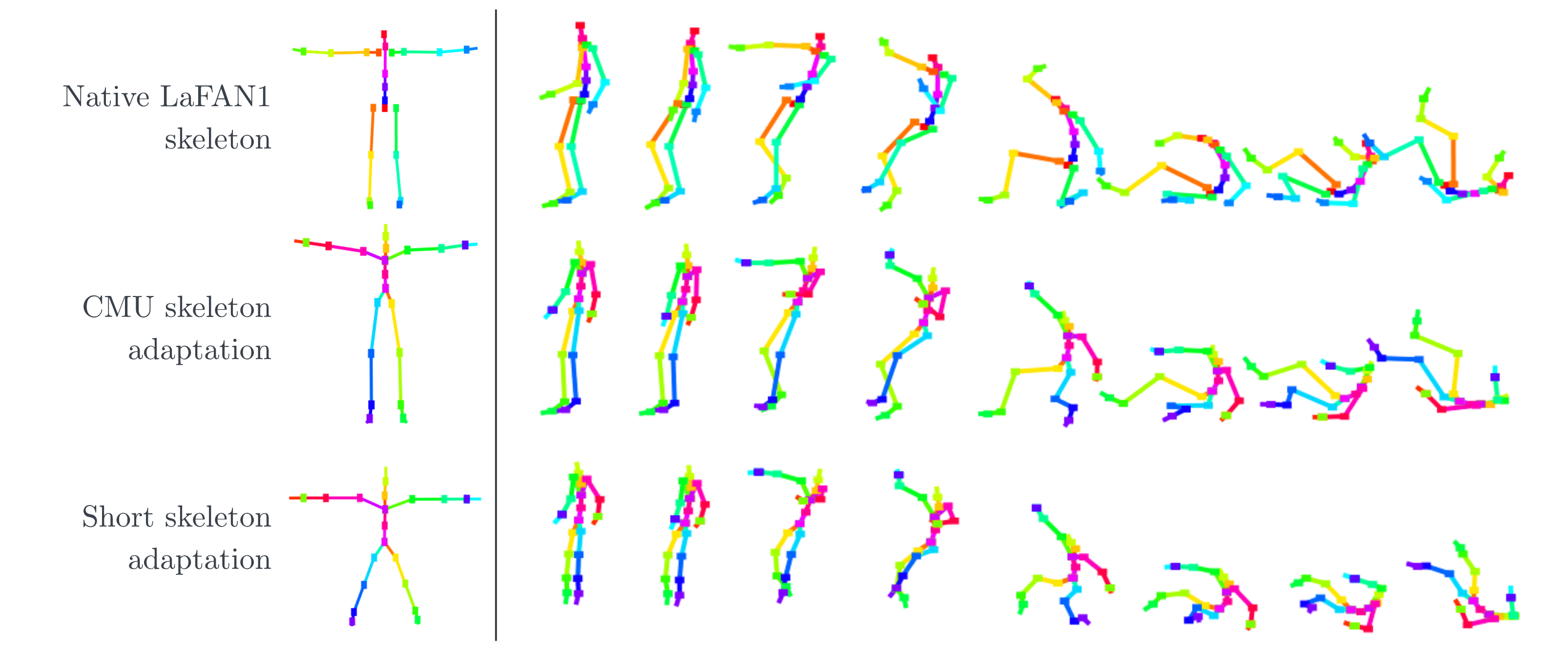}
    \caption{Re-targeted falling motion from LaFAN1 (top) on the CMU skeleton (middle) and a short custom skeleton (bottom), performed by our point cloud-to-motion model.}
    \label{fig:dwarf}
\end{figure}

Table \ref{tab:retarget} indicates the superior performance of our method over the state-of-the-art PS method in terms of visual similarity. At low $\sigma$ values, the higher relative $\mathcal{L}_{\text{KNN}}$ of the primal skeleton method underscores its geometric deviations from the original motion. In contrast, our method demonstrates significantly better visual adherence to the original geometry. 
As $\sigma$ increases, our method approaches near-perfect alignment with the original motion. Figure \ref{fig:retarget} visually indicates this, showing our method follows the original motion closely, while the latent consistency-based primal skeleton method struggles to generalise and decode the skeleton's internal structures without directly supervised objectives, such as the end-effector positions provided during training. In addition, Fig \ref{fig:dwarf} demonstrates that, like existing re-targeting approaches, our method is able to produce adequate results on disproportionate skeletons.

Since our approach is largely focused on enabling non-native motion interpolation supervision, we additionally compare our PC-MRL method against a CITL model supervised by PS in Table \ref{tab:interp_ps}. Due to the suboptimal re-targeting performance of PS, such interpolation models exhibit expectedly poor accuracy. 

\begin{table}
    \centering
    \resizebox{\textwidth}{!}{
        \setlength\tabcolsep{5pt}
        \begin{tabular}{|l|c c c|c c c|c c c|}
            \hline
            \multirow{2}{*}{\textbf{Method}} & \multicolumn{3}{|c|}{\textbf{L2P}$\downarrow$} & \multicolumn{3}{|c|}{\textbf{L2Q}$\downarrow$} & \multicolumn{3}{|c|}{\textbf{NPSS}$\downarrow$} \\
            & 5 & 15 & 30 & 5 & 15 & 30 & 5 & 15 & 30 \\
            \hline
            CITL - PS & 0.2043 & 0.6332 & 1.1524 & 0.2318 & 0.7428 & 1.1442 & 0.1122 & 0.4276 & 1.4622 \\
            CITL - Native & 0.0495 & 0.2306 & 0.4898 & 0.0979 & 0.3604 & 0.6609 & 0.1149 & 0.4684 & 1.1591 \\
            PC-MRL & 0.0494 & 0.2337 & 0.4960 & 0.1030 & 0.3771 & 0.6813 & 0.1370 & 0.5433 & 1.1760 \\
            \hline
        \end{tabular}
    }
    \caption{Average L2P, L2Q, and NPSS performance of CITL trained on native and PS-generated datasets, compared to our method.}
    \label{tab:interp_ps}
\end{table}

\section{Limitations}

Due to the inherent limitations of the point cloud representation, certain constraints on the applicability of our point cloud-to-motion method are inevitable. 
Primarily, the reliance of our point cloud re-targeting method on the relative nature of First-frame Offset Quaternions (FOQ) for most motion learning tasks necessitates the presence of at least \textit{one known pose}. In the absence of a known pose, relative quaternions are incapable of reconstructing absolute rotational values, which are essential for motion applications.
Secondly, due to our method of sampling point clouds with a consistent deviation from each bone segment, the resultant representation inevitably obscures finer details. This includes elements like fingers and facial controls, to an extent that is beyond rectification.

Our proposed objective function relies entirely on geometric segments, necessitating that all bones possess a non-zero length. This stipulation is deemed a reasonable prerequisite for generic human skeletons, as each bone is typically responsible for manipulating a discernible segment of the human body. Due to technical considerations, the skeletal configurations in each of our datasets included at least one bone of zero length, which we ignored in our experiments in favour of global quaternion rotations of all bones with non-zero length.

As a result of these limitations, we refrained from claiming state-of-the-art performance for general motion re-targeting tasks.

\section{Conclusion}
This paper introduces a novel learning-based motion interpolation method designed to enable cross-skeleton compatibility with existing motion datasets. Our proposed method PC-MRL employs a process of point cloud obfuscation and skeleton reconstruction. The point cloud space represents human motions in a non-hierarchic and skeleton-agnostic manner. It enables a KNN-based objective to be optimised without dataset supervision, guiding a neural network to generate high-quality motion features with any target skeleton. 
We address the rotational information loss of our point cloud format by presenting an offset quaternion strategy compatible with concurrent transformer-based models. 
Through extensive experiments, we have demonstrated the efficacy of PC-MRL in performing motion interpolation without relying on native motion data. Moreover, PC-MRL has achieved superior visual similarity metrics in the domain of motion re-targeting. We concluded our work by discussing the limitations of PC-MRL.

%
%
\bibliographystyle{splncs04}

\end{document}